\documentclass[10pt,twocolumn,letterpaper]{article}

\usepackage{cvpr}
\usepackage{times}
\usepackage{epsfig}
\usepackage{graphicx}
\usepackage{amsmath}
\usepackage{amssymb}

\usepackage[pagebackref=true,breaklinks=true,letterpaper=true,colorlinks,bookmarks=false]{hyperref}

\cvprfinalcopy 

\def\cvprPaperID{1923} 

\ifcvprfinal\fi
\begin{document}


\title{Knowledge Concentration: Learning 100K Object Classifiers in a Single CNN}

\author{Jiyang Gao$^{1}$ \quad Zijian (James) Guo$^{2}$ \quad Zhen Li$^2$  \quad Ram Nevatia$^1$ \\
$^1$University of Southern California \qquad $^2$Google Research \\
{\tt\small jiyangga@usc.edu, \{guozj, zhenli\}@google.com,\quad nevatia@usc.edu} 
}

\maketitle

\begin{abstract}
Fine-grained image labels are desirable for many computer vision applications, such as visual search or mobile AI assistant. These applications rely on image classification models that can produce hundreds of thousands (\emph{e.g.} 100K) of diversified fine-grained image labels on input images. However, training a network at this vocabulary scale is challenging, and suffers from intolerable large model size and slow training speed, which leads to unsatisfying classification performance. A straightforward solution would be training separate expert networks (specialists), with each specialist focusing on learning one specific vertical (\emph{e.g.} cars, birds...). However, deploying dozens of expert networks in a practical system would significantly increase system complexity and inference latency, and consumes large amounts of computational resources. To address these challenges, we propose a Knowledge Concentration method, which effectively transfers the knowledge from dozens of specialists (multiple teacher networks) into one single model (one student network) to classify 100K object categories. There are three salient aspects in our method: (1) a multi-teacher single-student knowledge distillation framework; (2) a self-paced learning mechanism to allow the student to learn from different teachers at various paces; (3) structurally connected layers to expand the student network capacity with limited extra parameters. We validate our method on OpenImage and a newly collected dataset, Entity-Foto-Tree (EFT), with 100K categories, and show that the proposed model performs significantly better than the baseline generalist model.
\end{abstract}

\section{Introduction}

Convolutional Neural Network (CNN) models has been rapidly improved in recent years, from AlexNet \cite{krizhevsky2012imagenet}, VGG \cite{simonyan2014very}, Inception \cite{szegedy2016rethinking}, ResNet \cite{he2016deep}, to DenseNet \cite{Huang_2017_CVPR}. When applied in 1000 category classification tasks in ImageNet \cite{deng2009imagenet} competition, these models have approached or exceeded human performance. However, in many computer vision applications, such as a mobile AI assistant, 1000 category labels are far from sufficient. After all, how often do the users need their phones to tell them that a coffee mug is a coffee mug? Users need the visual search system to tell them the name of a flower when they see a beautiful flower in a park, or tell them where to buy the items when they see fashionable items on street. To make computer vision systems practically useful in these applications, a wide varieties of fine-grained and informative labels are required.


\begin{figure}[t]
  \centering
    \includegraphics[width=0.48\textwidth]{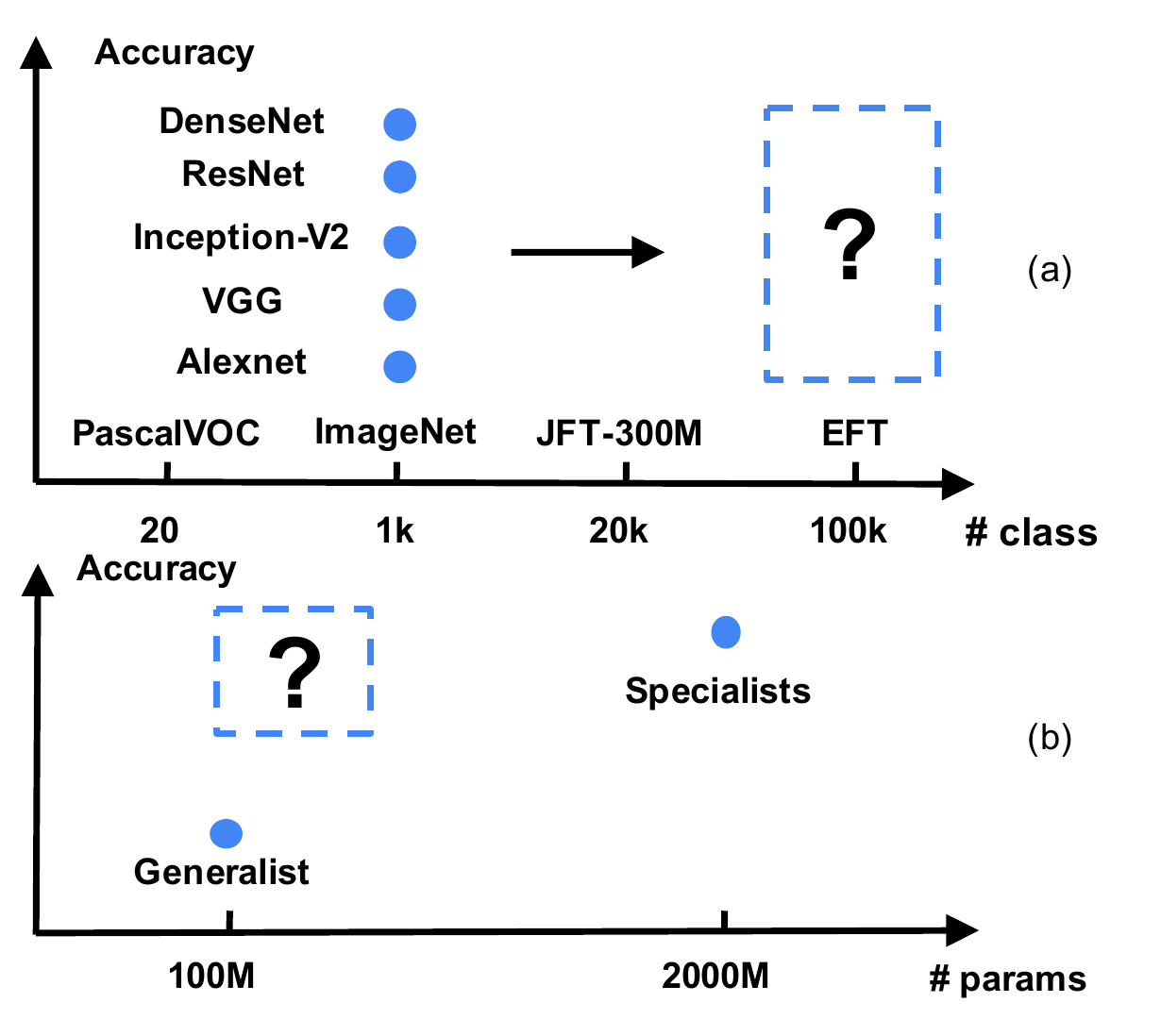}
    \caption{(a) Recent CNN architectures are designed and tested on 1000-class ImageNet dataset, what will happen when we apply them on much larger classification problems, \eg 100K classes? (b) Single generalist model (single model handles all classes) for 100K classes generates poor performance due to limited model capacity; multiple specialist models (one export for a subset of classes, \eg 20 subsets) consume large amount of computational resources and increase system complexity. Is there a method that use reasonable resource and also provide good performance?}
      \label{fig:problem}
    \vspace{-5pt}
\end{figure}

This paper tackles the challenge of training CNN with a vocabulary on the scale of O(100K), 100 times larger than the 1K labels in ImageNet \cite{deng2009imagenet} and 5 times larger than the 18K labels in the largest known dataset, JFT-300M \cite{Sun_2017_ICCV}, as shown in Figure \ref{fig:problem}. The challenges mainly lie in two aspects: limited CNN model capacity and slow training speed. The size (number of parameters) of CNN models is limited by GPU memory. On one hand, differentiating larger number of output classes requires stronger image representations and thus a larger bottleneck layer. On the other hand, large number of output classes limits the size of bottleneck layer because of limited GPU memory. For example, with a 4096-d bottleneck layer and 100K output classes, there are more than $4096 \times 100 \mbox{K} > 400\mbox{M}$ parameters in this fully connected layer. Together with parameters in other layers in the model, activations, and variable replicas required by optimizer, the training job would not fit in GPU memory. As for training speed, training models on the JFT-300M \cite{Sun_2017_ICCV} dataset for 1 epoch needed 2 weeks on 50 Nvidia K80 GPUs and the full training schedule takes around 10 epochs \cite{Sun_2017_ICCV}. Directly training model on 100K categories, with 5 times larger vocabulary, would consume prohibitively large amount of computational resources and much longer training time. 

One straightforward solution to solve the problems is to divide the vocabulary into semantically related \emph{subgroups} (or \emph{verticals}), \emph{e.g.}, grouping all fine-grained birds into one vertical while grouping all fine-grained cars into another vertical. We then train dozens of specialist networks, with each specialist network learns to classify limited number of categories in one vertical. Multiple specialist networks can be trained in parallel, which saves wall time of training. However, deploying dozens of separate fine-grained models in a practical system would increase system complexity, inference latency, and consume large amount of computational resources, as shown in Figure \ref{fig:problem}.

To address these challenges, we propose a multi-teacher single-student knowledge concentration method, which effectively merges the knowledge from dozens of teachers (\emph{i.e.} specialists) into a single student model. To better transfer the knowledge from multiple teachers to a student, we design a self-paced learning mechanism to allow the student to learn from different teachers at various paces. The intuition of our work come from daily experience: (1) a student absorbs knowledge from multiple teachers at school on different course subjects, and (2) a good student spends different efforts on different course subjects, since they have different amounts of material to learn. To expand the student network capacity but limiting the of parameters, we explore and design structurally connected layers, which includes generic units for vertical agnostic representation and individual units for intra-vertical representations. We perform experiments on a newly collected dataset, Entity-Foto-Tree (EFT), with 100K categories and on OpenImage. The EFT dataset are organized in a hierarchy and clustered into 20 verticals. Experimental results show that the proposed knowledge concentration method outperforms the baseline models by a large margin on both datasets.

Our contributions are three-fold:

(1) We design a novel multi-teacher single-student knowledge distillation method to transfer knowledge from the specialists to the generalist, and a self-paced learning mechanism allowing the student to learn at different paces from different teachers. 

(2) We design and explore different types of structurally connected layers to expand network capacity with limited number of parameters.

(3) We evaluate the proposed methods on EFT and OpenImage datasets, and show significant performance improvements.

\section{Related Work}

\textbf{Knowledge distillation.} The concept of knowledge distillation is originally proposed by Hinton \etal \cite{hinton2015distilling}, which uses the soft targets generated by a bigger and deeper network to train a smaller and shallower network and achieves similar performance as the deeper network. Ba \etal \cite{ba2014deep} also demonstrated that shallow feed-forward nets can learn the complex functions previously learned by deep nets. Romero \etal \cite{romero2014fitnets} extended Hinton's work\cite{hinton2015distilling} by using not only the outputs but also the intermediate representations learned by the teacher as hints to train the student network. The aforementioned methods fall into single-teacher single-student manner, where the task of the teacher and the student is the same. Rusu \etal \cite{rusu2015policy}  proposed a multi-teacher single-student policy distillation method that can distill multiple policies of reinforcement learning agents to a single network. Our Knowledge Concentration method is also a multi-teacher single-student framework.

\textbf{Transfer learning.}
Fine-tuning is a common strategy in transfer learning with neural networks \cite{sharif2014cnn, yosinski2014transferable, azizpour2016factors, oquab2014learning}. Oquab \etal \cite{oquab2014learning} showed that CNN image representations learned from large-scale annotated datasets can be efficiently transferred to other tasks with limited amount of training data by fine-tuning the network. Li \etal \cite{li2016learning} proposed a Learning without Forgetting method, which uses only new task data to train the network while preserving the original capabilities. Wang \etal \cite{Wang_2017_CVPR} demonstrate that ``growing" a CNN with additional units for fine-tuning on a new task, newly-added units should be appropriately normalized to allow for a pace of learning that is consistent with existing units. Our self-paced learning mechanism is related to \cite{Wang_2017_CVPR}, we propose a vertical level pace adjustment mechanism and a dynamic pace initialization method, while \cite{Wang_2017_CVPR} is on node level with a fixed initialization.

\textbf{Branched CNN.} There is some previous work on branched (\emph{i.e.} tree-structured) CNN, which is related to our structural connected layers. Ahmed \etal \cite{ahmed2016network} presented a tree-structured network which contains multiple branches. The branches share a common base network and each branch classifies a subset of similar categories. Yan \etal \cite{Yan_2015_ICCV} introduced hierarchical deep CNNs by embedding deep CNNs into a category hierarchy, which separates easy classes using a coarse category classifier while distinguishing difficult classes using fine category classifiers. The branches in tree-structured CNN can be viewed as a special type of structurally connected layers.

\textbf{Fine-grained classification.} Fine-grained image classification requires the model to discern subtle differences among similar categories. Many methods are designed based on CNN models \cite{zhang2014part, lin2015bilinear, sharif2014cnn} that learn feature representations from data directly for classification. Cui \etal \cite{cui2016fine} proposed a generic iterative framework for fine-grained categorization and dataset bootstrapping which uses deep metric learning with humans in the loop. Zhou \etal \cite{zhou2016fine} exploited the rich relationships through bipartite-graph labels. Fu \etal \cite{Fu_2017_CVPR} designed a recurrent attention CNN which recursively learns discriminative region features to discern subtle differences for similar classes. He \etal \cite{He_2017_CVPR} proposed a two-stream model that combines the vision and language for learning latent semantic representations for fine-grained classification.

\section{Knowledge Concentration}
In this section, we introduce Knowledge Concentration method in three parts: (1) multi-teacher single-student knowledge distillation framework; (2) structurally connected layers and (3) self-paced learning mechanism.
\subsection{Multi-teacher Single-student Knowledge Distillation}
The original idea of knowledge distillation proposed by Hinton \emph{et al.} \cite{hinton2015distilling} focuses on distilling knowledge from a single teacher to a single student; we present a framework for integrating knowledge from multiple teachers to a single student. 

\textbf{Teacher model training.} In our framework, teachers are a set of trained specialist models. Each specialist model is a Convolutional Neural Network (CNN) trained on a \emph{vertical} $V_i$. The classes in one vertical are semantically related and usually visually confusing, e.g fine-grained cars are in one vertical, fine-grained birds are in another vertical. We use $T(V_i)$ to represent the specialist model for vertical $V_i$. In a large scale dataset, such as JFT-300M \cite{Sun_2017_ICCV}, categories are usually organized in a semantically hierarchical structure, and thus not all classes are mutually exclusive. Therefore, we use sigmoid cross-entropy loss instead of softmax cross-entropy to train the teacher models.

\begin{equation}
    L_{t}=-\frac{1}{N_b}\sum_{i=1}^{N_b}\sum_{j=1}^{N_c}(log(\sigma(x_{ij}))z_{ij}+log(1-\sigma{(x_{ij})})(1-z_{ij}))
\end{equation}

where $N_b$ is the batch size, $N_c$ is the number of classes, $z_{ij}$ and $x_{ij}$ are the label and output logit of sample $i$ for class $j$ respectively, $\sigma(.)$ is the sigmoid function. If the dataset organizes the categories in a hierarchical structure, then the labels are dynamically propagated to ancestor nodes, \eg  a training sample of ``golden retriever" would also be a positive sample for ``dog", ``mammal" and ``animal", as ``blackbird" is a sub-type of ``bird".  We call this label propagation process as ``label smearing". 

\begin{figure}[h]
  \centering
    \includegraphics[width=0.48\textwidth]{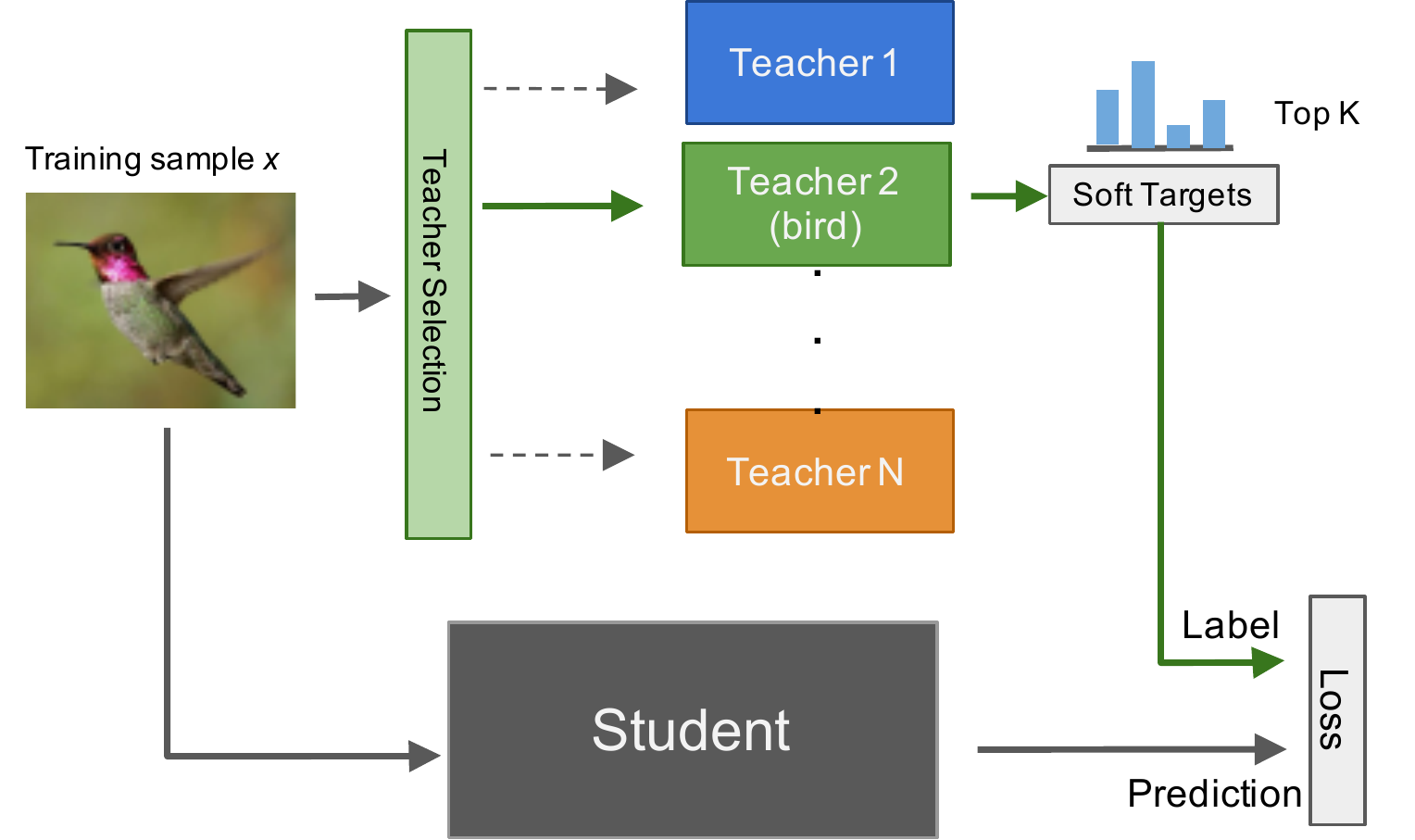}
    \caption{Multi-teacher single-student knowledge distillation framework: training samples are input to the corresponding specialist model, the output probability distributions are used as soft targets to optimize the student network.}
      \label{fig:teacher}
\end{figure}

\textbf{Knowledge distillation from multiple teachers.} Given a training sample $s$ and the groundtruth label $z$, instead of using it for training directly, we first generate a probability distribution $p$ by feeding $s$ into the corresponding teacher (\emph{i.e.} specialist) model, as shown in Figure \ref{fig:teacher}. The teacher model is selected by a vertical mapping function $F_{map}$, which maps a label $z$ to a vertical $V_i$. The generated probability distribution is $p=G(T(F_{map}(z)),s)$, where $G$ is the mapping from the image space to the label space of a teacher model. For each sample, we keep the top $K$ labels and use their probabilities as soft targets $z^*$ for distillation. For the categories that are not in the top $K$ categories of the vertical or that are not in the selected vertical $V_i$, we set their labels as $0$ (\emph{i.e.} negative). We use sigmoid cross entropy loss to train the student network, which is the same for teacher network. 

\begin{equation}
    L_{s}=-\frac{1}{N_b}\sum_{i=1}^{N_b}\sum_{j=1}^{N_c}(log(\sigma(x_{ij}))z^*_{ij}+log(1-\sigma{(x_{ij})})(1-z^*_{ij}))
\end{equation}
where $z^*_{ij}$ and $x_{ij}$ are the soft target and output logit of sample $i$ for class $j$ respectively. Note that, label smearing is not used for student network training. 

\subsection{Structurally Connected Layers}
With large number of output classes, the number of parameters in the last Fully Connected (FC) layer can be prohibitively large. Therefore, shrinking the size of this bottleneck layer is essential if using an FC layer structure. In order to expand the model capacity efficiently without blowing GPU memory, we design Structurally Connected (SC) layers. Unlike densely connected FC layers, SC layers sparsely connect the consecutive layers and thus significantly reduce the number of parameters. We consider the CNN architecture as ``base network + 2 top layers". The base network contains the convolutional layers and the top layers can be FC or SC layers. We denote the first top layer (\emph{i.e.} the layer connecting to the base network) as top-1 layer and second top layer as top-2 layer (\emph{i.e.} the layer to top of top-1 layer and connecting the logit outputs). 

\begin{figure*}[h]
  \centering
    \includegraphics[width=0.96\textwidth]{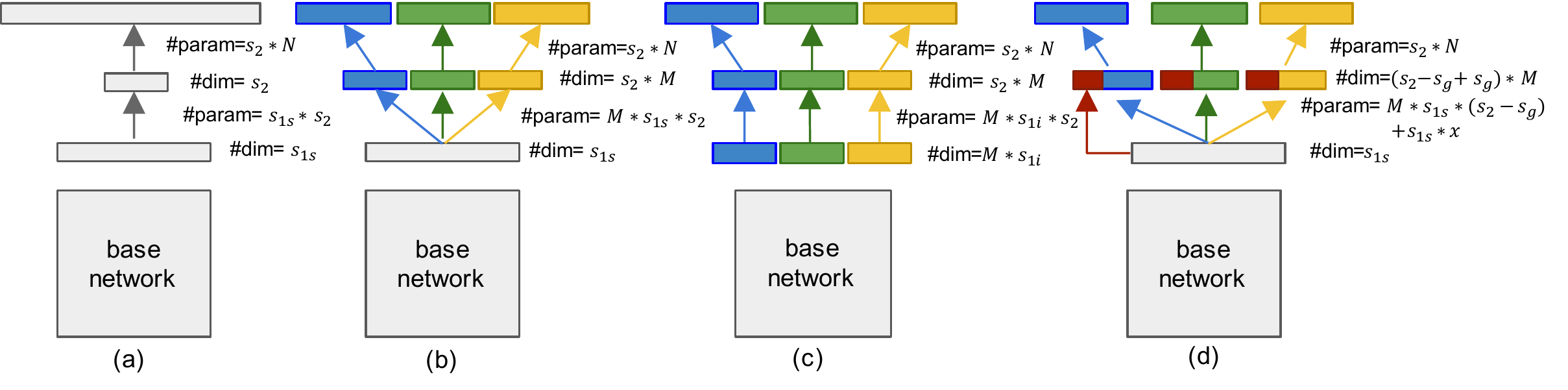}
    \caption{A regular Fully Connected (FC) layer structure is shown in (a); three types of Structurally Connected (SC) layer structures are shown in (b),(c),(d), which are  ``FC($s_{1s}$)-SC($s_{2}$)", ``SC($s_{1i}$)-SC($s_{2}$)" and ``FC($s_{1s}$)-SC($s_{2},s_g$)" respectively. The layer dimension and number of parameters are shown.}
      \label{fig:sclayers}
\end{figure*}

Figure \ref{fig:sclayers} (a) shows the architecture employing only the fully connected layers; it is named as ``FC($s_1$)-FC($s_2$)", where $s_1$ is size for top-1 layer and $s_2$ is the size for top-2 layer. The total number of parameters in the top layers is $s_2\times N+s_2\times s_1+s_1\times s_b$, where $N$ is the number of output classes, and $s_b$ is the output layer size of the base network. Figure \ref{fig:sclayers} (b) (c) (d) shows three types of network architectures employing different combinations of FC and SC layers. 

Figure \ref{fig:sclayers} (b) shows the network architecture where verticals have separate top-2 layers and shared top-1 layer. The total number of parameters in the top layers is $s_2\times N+M\times s_2\times s_{1s}+s_{1s}\times s_b$, where $M$ is the number of verticals and $s_{1s}$ is the size of shared top-1 layer. We name this type of architecture ``FC($s_{1s}$)-SC($s_{2}$)". Figure \ref{fig:sclayers} (c) shows the second type of network architecture, ``SC($s_{1i}$)-SC($s_{2}$)", where verticals have not only separate top-2 layers but also separate top-1 layers. All the top-1 layers share the same size $s_{1i}$. The total number of parameters in the top layers is $s_2\times N+M\times s_2\times s_{1i}+M\times s_{1i}\times s_b$. Figure \ref{fig:sclayers} (d) shows the third type, where we split the top-2 layers into a generic part and individual parts; the generic part is shared in all top-2 layers. As we restrict the size of the top-2 layer, the size of individual part is decreased to $s_2-x$, where $x$ is the size of the generic part. This type is called ``FC($s_{1s}$)-SC($s_{2},x$)". The total number of parameters in the top layers is $s_2\times N+M\times (s_2-s_g)\times s_{1s}+x\times s_{1s}+s_{1s}\times s_b$. Actually, ``FC($s_{1s}$)-SC($s_{2}$)" can be viewed as a special case for ``FC($s_{1s}$)-SC($s_{2},s_g$)", where $s_g$ (generic part) is zero.

For fair comparison, we set the top-2 layer of (a), (b), (c) and (d) to have the same size, $s_2$, since the number of parameters connecting the logit layer and the top-2 layer dominate the total number of parameters. 

\subsection{Self-paced Learning for Different Verticals}
\begin{figure}[h]
  \centering
    \includegraphics[width=0.30\textwidth]{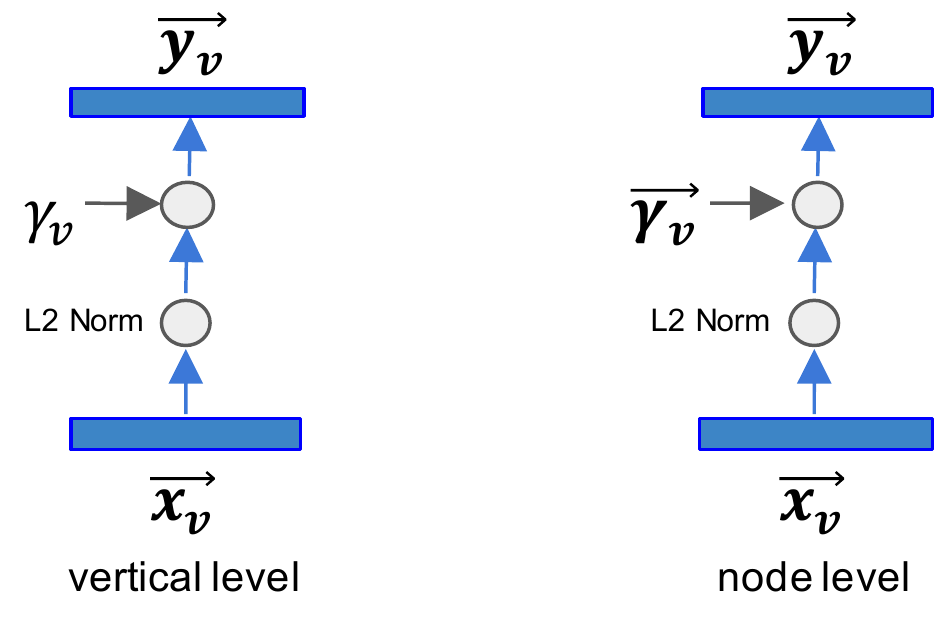}
    \caption{Self-paced learning mechanism by logits L2-normalization and multiplication with trainable scaling factors on either vertical-level or node-level.}
      \label{fig:selfpaced}
\end{figure}

Different verticals have different number of classes and different amounts of training data. Therefore, when transferring knowledge from multiple specialists into the student network, it is desirable to control learning from different verticals at different paces. We design a mechanism that allows each vertical to learn at its own pace. As shown in Figure \ref{fig:selfpaced}, the logits are first L2-normalized inside each vertical and then multiplied by scaling factors. The scaling factors $\gamma$ can be applied either on the vertical level or on the class level. Vertical level scaling factor is assigned to each vertical, the final output logits is $\vec{y_v}=\gamma_v||\vec{x_v}||$, where $||x||$ means L2-normalization, $\vec{x_v}$ is the original logits of vertical $v$, $\gamma_v$ is a single scaling factor applied on $v$, and $\vec{y_v}$ is the output vector. Class level scaling factor is assigned to each output node (\emph{i.e.} class), the final output logits is $\vec{y_v}=\vec{\gamma_v}||\vec{x_v}||$, where $\gamma_v$ is the scaling factor vector for vertical $v$, each class inside this vertical has a individual scaling factor, $\vec{y_v}$ is the element-wise multiplication between $\vec{\gamma_v}$ and $||\vec{x_v}||$.

The scaling factors $\gamma$ are trainable variables. We consider the gradient deduction for a node (\emph{i.e.} class) $i$ after L2-normalization. We use $\hat{\vec{x_i}}$ as the logit after normalization for node $i$.
\begin{equation}
    \hat{x_i}=||\vec{x}||_i 
\end{equation}
After multiplying the scaling factor $\gamma_i$, the logit $y_i$ is
\begin{equation}
    y_i=\gamma_i \hat{x_i}
\end{equation}
For class-level scaling factor, $\gamma_i$ is different for different classes; for vertical-level scaling factor, $\gamma_i$ is the same for the classes in one vertical. In back-propagation, the gradient for $x_i$ is 
\begin{equation}
\begin{split}
    \frac{\partial{l}}{\partial{x_i}}=\frac{\partial{l}}{\partial{\hat{x_i}}} \frac{\partial{\hat{x_i}}}{\partial{x_i}}=\gamma_i\frac{\partial{l}}{\partial{y_i}}(\frac{1}{||x||_2}-\frac{x_i^2}{||x||_2^3})
\end{split}
\end{equation}
We can see that larger $\gamma_i$ will produce larger logits and also larger gradients, thus $\gamma_i$ adjusts the learning pace for class $i$.

L2 normalization would make the logit $\hat{x_i}$ much smaller than before, and thus the network is hard to train. As derived above, the ratio between $\partial{\hat{x_i}}$ and $\partial{x_i}$ is
\begin{equation}
\begin{split}
    \frac{\partial{\hat{x_j}}}{\partial{x_j}}&=\frac{1}{\sqrt{\sum{x_j^2}}}(1-\frac{x_j^2}{\sum{x}_j^2})\\&\approx\frac{1}{\sqrt{\sum{x_j^2}}}\approx\frac{1}{\sqrt{N_v}\sqrt{E(x_j^2)}}\propto\frac{1}{\sqrt{N}}
\end{split}
\end{equation}
where $N_v$ is the number of classes in vertical $v$, which contains class $i$. As all parameters are initialized randomly, $\frac{x_j^2}{\sum{x}_j^2}$ is close to $0$. $E(x_j^2)$ is the expectation of $x_j^2$. We can see that the ratio between is $\partial{\hat{x_i}}$ and $\partial{x_i}$ is proportional to $\frac{1}{\sqrt{N_v}}$. Thus, we use $\sqrt{N_v}$ as the initialization value for $\gamma_i$ to recover the gradient scale.

\section{Evaluation}
In this section, we introduce the evaluation metrics and experimental setup, and discuss the experiments results.

\subsection{Experiments on EFT}
\textbf{Dataset.} The Entity-Foto-Tree (EFT) dataset contains 400 million images on 100 thousand classes. The class labels are physical entities organized in a tree-like hierarchy, which contains 20 diversified verticals: \textit{aircraft, bird, car, dish, drink, gadget, hardware, home and garden, house, human settlement, infrastructure, ingredient, invertebrate, landmark, mammal, sports, watercraft, weapon, wildlife, plant}. The EFT test set contains 5 million images, with 50 held out images for each of the 100K classes.

\textbf{Base network.} The base network is a modified Inception-V2 \cite{szegedy2016rethinking}: instead of mean-pooling the ``mixed5c" layer to 1024-dimensional, we first mean-pool "mixed5c" with a $3\times 3$ kernel to a $3\times 3\times 1024$ feature map and then reshape it to 9216 dimensional feature. All models tested on EFT use the same base network.

\textbf{Baselines.} The first baseline method is a generalist CNN that is directly trained to classify all 100K classes; For the baseline generalist CNN, we don't use the proposed training method on it, but just use the standard training procedure. The architecture shown in Figure \ref{fig:sclayers} (a) is adopted, the sizes of top-1 and top-2 layers are 4096 and 512 respectively. We name it ``FC4096-FC512". This method serves as a performance lower bound. The second baseline method is to train separate specialist models for each of the 20 verticals. For the specialist models, both the sizes of top-1 and top-2 layers are set to 4096 to maximize the model capacity, and we name this architecture as `FC4096-FC4096". These specialist models serve as the performance upper bound.

\textbf{Experiment setup.} We set $K$ (the number of selected top categories for training student in knowledge distillation) to 100. We train the models in Tensorflow with 50 GPU workers (Nvidia P100) and 25 parameter servers. Adagrad \cite{duchi2011adaptive} is used to optimized the model, learning rate is set to 0.001, Batch size is set to 64. Each model is trained for around 5 epochs, which takes approximately 40 days.

\textbf{Evaluation metric.} To fairly compare the performance of the proposed methods with the baseline generalist and the specialist models, we compare the performance on each vertical. We use (mean) per-vertical average precision as evaluation metric. Specifically, given a vertical, for a generalist model, we first prune the output labels by the label list of the vertical and then calculate the mean average precision on this vertical, which is called per-vertical average precision (pvap). Mean per-vertical average precision (mpvap) is the average of all per-vertical average precision.

\textbf{Evaluation on Multi-teacher Single-student Distillation.} We compare the generalist models trained without multi-teacher distillation and with multi-teacher distillation. Two network architectures are tested: (1) FC4096-FC512, which is used in baseline generalist, shown in Figure \ref{fig:sclayers} (a); (2) SC(512)-SC(512), the second type of SC layers introduced in Section 3.2, we set $s_{1s}=512, s_2=512$. The results are shown in Table \ref{table:teacher}. We can see that for both architectures, using multi-teacher distillation improves the performance. Specifically, multi-teacher distillation improves the performance from 27.7 to 31.5 for FC4096-FC512 and from 27.6 to 31.9 for SC512-SC512.

\begin{table}[h]
\centering
\caption{Comparison of multi-teacher single-student knowledge distillation on EFT dataset, mean per-vertical average precision (\%) is reported.}
\label{table:teacher}
\begin{tabular}{l|c|c}
\hline
    method     & w/o distillation & w/ distillation \\ \hline
FC4096-FC512 & 27.7             & \textbf{31.5}            \\ \hline
SC512-SC512  & 27.6             & \textbf{31.9}            \\ \hline
\end{tabular}
\vspace{-5pt}
\end{table}

\textbf{Comparison on different SC layers.} We evaluate three types of SC layer architectures shown in Figure \ref{fig:sclayers} (b)(c)(d). We fix top-2 layer size to be 512 for fair comparison. Specifically, we first compare FC4096-FC512, FC4096-SC512 and SC512-SC512 to see the depth impact of SC layers. All three models are trained with multi-teacher distillation. As shown in Table \ref{table:sc1}, FC4096-SC512 achieves the best performance. Specifically, comparing FC4096-FC512 and FC4096-SC512, results show that FC4096-SC512 expand the model capacity efficiently and improves the performance from 31.5 to 32.9. The FC4096-SC512 structure expands the model capacity by allowing each vertical to have its own bottleneck layer and increases the total bottleneck layer size to $512\times N$. At the same time, it removes the cross-vertical connections and successfully avoids increasing the number of the parameters in the classification layer. Here, removing the cross-vertical connections does not hurt model performance because the top-2 layer is semantically strong and the verticals are semantically separated.

Comparing FC4096-SC512 and SC512-SC512, we can see that although SC512-SC512 contains more number of parameters, the performance of it is lower than FC4096-SC512, which shows that it is important to share the lower level features (top-1 layer) among verticals.

\begin{table}[h]
\centering
\caption{Comparison of different SC layer architectures on EFT dataset for classification performance (mpvap \%) and model size (number of parameters).}
\label{table:sc1}
\begin{tabular}{l|cc}
\hline
method    & mpvap(\%) & \#param \\ \hline
FC4096-FC512 & 31.5  & 98M     \\ \hline
FC4096-SC512 & \textbf{32.9}  & 135M    \\ \hline
SC512-SC512  & 31.9  & 155M  \\ \hline 
\end{tabular}
\vspace{-5pt}
\end{table}

We also test the third type of SC layer structure shown in Figure \ref{fig:sclayers} (d) by varying the size of generic part $x$. We set $x=0,128,256,384,512$, and the results are shown in Table \ref{table:sc2}. It can be seen that generic part on top-2 layer is not helpful for classification performance. The reason is that the generic part reduces total top-2 layer size and makes top-2 layer the bottleneck of information flow. 

\begin{table}[h]
\centering
\caption{Comparison of generic feature size of 0, 128, 256, 384 and 512 on EFT dataset, mean per-vertical average precision (mpvap \%) is reported. }
\label{table:sc2}
\begin{tabular}{l|cc}
\hline
 method    & mpvap(\%) & \#param \\ \hline
FC4096-SC(512,0)   & \textbf{32.9}  & 135M    \\ \hline
FC4096-SC(512,128) & 32.8  & 127M    \\ \hline
FC4096-SC(512,256) & 32.4  & 117M    \\ \hline
FC4096-SC(512,384) & 31.4  & 108M    \\ \hline
FC4096-SC(512,512) & 31.5  & 98M    \\ \hline
\end{tabular}
\end{table}
\begin{table*}[]

    \begin{center}
    \caption{Comparison of different methods on per-vertical average precision on EFT. In the method column,  ``+D" means using multi-teacher distillation, ``+S" means using self-paced learning. The specialist architecture is FC4096-FC4096, totally 20 of them. Abbreviation is used to fit the width of the paper: ``acraft"=aircraft, ``hdwr"=hardware, ``hm-gd"=home\&gardon, ``hm-st"=human settlement, ``infra"=infrastructure, ``igrdt"=ingredient, ``invbt"=invertebrate, ``ldmk"=landmark, ``mamal"=mammal, ``wcraft"=watercraft, ``wdlf"=wildlife}
    \resizebox{\textwidth}{!}{
    \begin{tabular}{l|cccccccccccccccccccc|c|c}
    \hline
    Method & acraft & bird & car & dish & drink & gadget & hdwr & hm-gd & house & hm-st & infra & igrdt & invbt & ldmk & mamal & sports & wcraft & weapon & wdlf & plant & mean & \#param \\ \hline\hline
    FC4096-FC512 & 22.7 & 33.4 & 51.8 & 21.5 & 30.8 & 38.4 & 28.6 & 34.6 & 26.6 & 13.6 & 17.5 & 26.8 & 25.9 & 17.2 & 29.1 & 39.2 & 23.6 & 25.9 & 15.6 & 31.0 & 27.7 & 98M\\\hline
    FC4096-FC512+D & 25.8 & 42.8 & 55.5 & 24.3 & 34.0 & 42.2 & 31.9 & 38.4 & 27.9 & 15.8 & 18.6 & 30.3 & 31.5 & 20.1 & 35.5 & 41.9 & 25.4 & 29.3 & 24.7 & 34.6 & 31.5 & 98M\\\hline
    FC4096-SC512+D & 27.5 & 44.7 & 55.9 & 26.9 & 35.2 & 44.7 & 32.9 & 39.3 & 28.6 & 16.7 & 19.2 & 32.1 & 33.3 & 22.8 & 36.4 & 42.5 & 26.1 & 30.3 & 26.0 & 36.4 & 32.9 & 135M\\\hline
    FC4096-SC512+D+S & \textbf{37.3} & \textbf{51.6} & \textbf{59.2} & \textbf{33.0} & \textbf{38.8} & \textbf{52.4} & \textbf{34.5} & \textbf{42.8} & \textbf{33.5} & \textbf{19.0} & \textbf{22.0} & \textbf{35.9} & \textbf{40.5} & \textbf{33.2} & \textbf{38.8} & \textbf{44.1} & \textbf{31.1} & \textbf{33.8} & \textbf{33.2} & \textbf{42.8} & \textbf{37.9} & 135M\\\hline
    Specialists & 67.9 & 82.9 & 74.2 &71.9& 72.9 & 82.4 & 59.6 & 76.6 & 78.6 & 45.9 & 60.5 & 66.4 & 71.5 & 79.1 & 60.9 & 66.0 & 57.7 & 51.4 & 59.3 & 75.1 & 68.0 & 2260M\\\hline

    \end{tabular}
    }
    \end{center}
    
    \label{table:detail}
    \vspace{-5pt}
\end{table*}

\begin{figure}[h]
  \centering
    \includegraphics[width=0.43\textwidth]{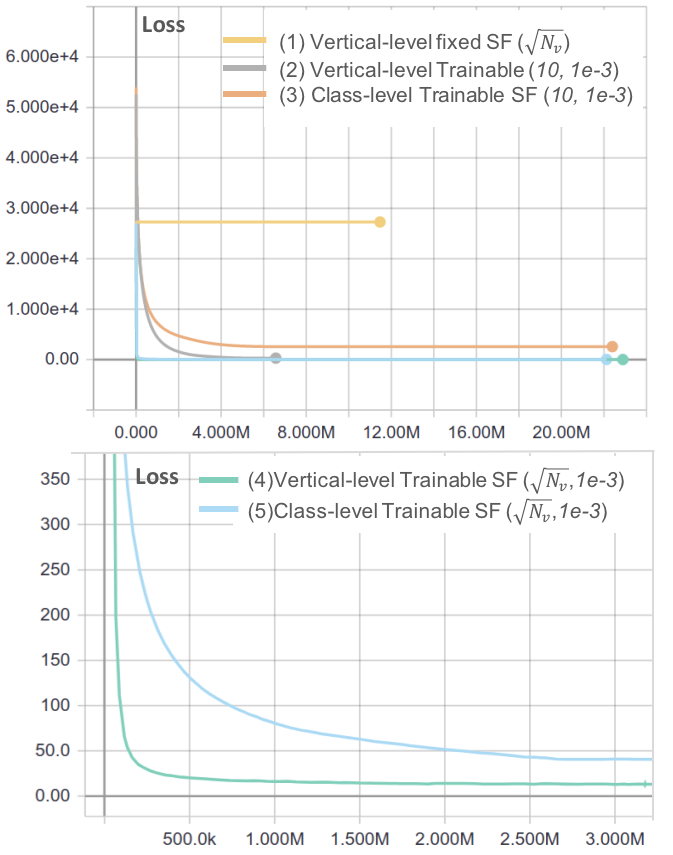}
    \caption{Loss comparison of five different system variants for self-paced learning mechanism. The lower figure is enlarged view of the upper figure, the lower figure shows the loss curve for ``trainable vertical-level scaling factors initialized by [$mean=\sqrt{N_v}$, $dev=1e-3$]" and ``trainable class-level scaling factors initialized by [$mean=\sqrt{N_v}$, $dev=1e-3$]".}
      \label{fig:sfloss1}
\end{figure}

\textbf{Evaluation of self-paced learning.} We first discuss the impact of initialization value of $\gamma$, and then compare the performance of vertical-level scaling factors and class-level scaling factors. The architecture we use is FC4096-SC512, as it achieves the best performance among all architectures. Multi-teacher distillation is adopted for all experiments. As shown in Figure \ref{fig:sfloss1}, we compare 5 settings of self-paced learning: (1) fixed vertical-level scaling factors initialized by $\sqrt{N_v}$; (2) trainable vertical-level scaling factors initialized by [$mean=10$, $dev=1e-3$]; (3) trainable class-level scaling factors initialized by [$mean=10$, $dev=1e-3$]; (4) trainable vertical-level scaling factors initialized by [$mean=\sqrt{N_v}$, $dev=1e-3$]; (5) trainable class-level scaling factors initialized by [$mean=\sqrt{N_v}$, $dev=1e-3$]. The loss optimization figures for all variants discussed above are shown in Figure \ref{fig:sfloss1}.

Comparing (1) and (4), although (1) and (4) are both initialized as $\sqrt{N_v}$ and applied on vertical level, fixed scaling factors couldn't optimize the model well, and scaling factors should be adjustable by different verticals. Comparing (2) and (3), (4) and (5), results show that vertical-level scaling factors generally work better than class-level scaling factors. The reason we believe is that class-level scaling factors introduce too many additional parameters, which in turn makes the optimization harder. Comparing (2) and (4), it can be seen that initializing the scaling factors $\sqrt{N_v}$ is critical for model optimization. The reason, as we discussed before, is that normalization changes the scale of logits and $\sqrt{N_v}$ provides a good estimation for the logit scale.

Overall, the optimization of (1) (2) and (3) are not converged, (4) and (5) are converged. Thus, we further compare the classification performance between (4) and (5), which is shown in Table \ref{table:spl}. We can see that class-level self-paced learning have almost no effect on the classification performance. Vertical-level self-paced learning increase the performance by a large margin, which shows the effectiveness for the proposed vertical-level self-paced learning. We think the reason is that class-level self-paced learning introduces too many additional parameters, which in turn make the optimization more difficult.

\begin{table}[h]
\centering
\caption{Comparison of self-paced learning on EFT dataset, mean per-vertical average precision (mpvap \%) is reported.}
\label{table:spl}
\begin{tabular}{l|c}
\hline
method       & mpvap(\%) \\ \hline
w/o self-paced learning            & 32.9  \\ \hline
w/ class-level     & 32.8  \\ \hline
w/ vertical-level & \textbf{37.9}  \\ \hline
\end{tabular}
\vspace{-5pt}
\end{table}

\begin{table*}[h!]
\centering
\caption{Comparison of different methods on per-vertical average precision on Openimage. We report (mean) per-vertical average precision (\%). In the method column, ``+D" means using multi-teacher distillation, ``+S" means using self-paced learning.}
\label{table:openimage}
\begin{tabular}{l|ccccc|c}
\hline
    Method                & food & organism & man made objects & mode of transport & others & mean \\ \hline
FC512-FC512              &  68.8 &	67.5	&50.5	&70.2	&52.9	& 62.0 \\ \hline
FC512-FC512+D          &  71.3 &	69.0	&51.8	&71.4	&52.7	& 63.2      \\ \hline
FC512-SC512+D            &  72.4&	69.5	&52.6	&71.9	&53.6	&64.0        \\ \hline
FC512-SC512+D+S            &  \textbf{75.0}	&\textbf{71.9}	&\textbf{54.6}	&\textbf{73.9}	&\textbf{53.9}	& \textbf{65.8}      \\ \hline 

5*FC512-FC512(Specialists) &  80.8	& 75.6	& 53.7	& 79.9	& 51.7	& 68.4     \\ \hline
\end{tabular}
\end{table*}

\textbf{Comparison on verticals.} In Table \ref{table:detail}, we list the per-vertical performance for 5 methods:(1) FC4096-FC512, this is the baseline generalist method, (2) FC4096-FC512 with multi-teacher distillation (FC4096-FC512+D), (3) FC4096-SC512 with multi-teacher distillation (FC4096-SC512+D), (4) FC4096-SC512 with multi-teacher distillation and vertical level self-paced learning (FC4096-SC512+D+S), (5) 20 specialist models, each model focuses on one vertical, the model architecture is FC4096-FC4096, distllation and self-paced learning are not used.  We can see that with the help of multi-teacher single-student distillation, structurally connected layers and self-paced learning, the performance is improved consistently for all verticals. Comparing the baseline generalist model and the FC4096-SC512+D+S, the performance is improved by 37\% on mpvap with only 37\% more parameters. Although 20 specialist models can achieve 109\% improvement, but the cost is 2260\% more parameters and 20 separate models, which consumes many more computational resources.

\begin{figure*}[h]
  \centering
    \includegraphics[width=0.96\textwidth]{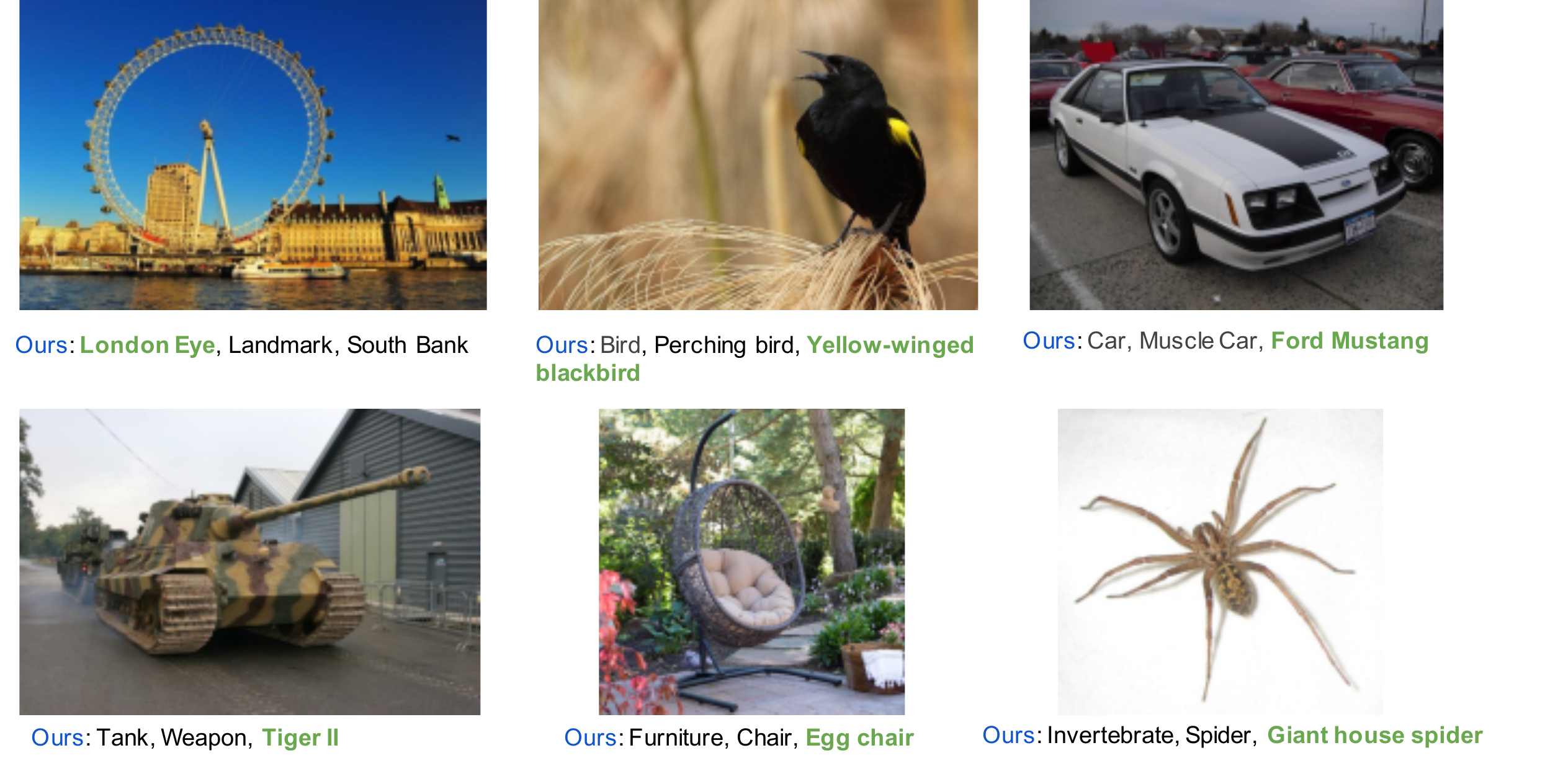}
    \caption{Visual examples on EFT dataset. Top-3 predictions are listed and the green color represents the correct prediction.}
      \label{fig:example}
\end{figure*}

\subsection{Experiments on OpenImage}
Openimage is a much smaller dataset compared with EFT dataset, and a normal deep CNN model \cite{szegedy2016rethinking} will not meet a model capacity problem. Therefore, we use a smaller and shallower network to simulate the gap between model capacity and scale of classes, such as the gap between Inception and EFT. The general CNN architecture includes a base network (smaller) and two top layers. 

\textbf{Dataset.} Open Images is a dataset of about 9 million images which have image-level labels and bounding boxes. The validation set contains another 170 thousand images. The total number of classes is around 6000 and the classes are organized in a hierarchical structure. We split all classes to 5 verticals: food, organism, man made objects, mode of transport and others.

\textbf{Base network.} The network should be smaller enough to show the model capacity restriction, so that the specialist model would perform better than generalist model on a certain vertical. Based on our experiments, we decided to use the ``Mixed\_1"+``Mixed\_2"+``Mixed\_3" of Inception-V2 as the base network. The ``Mixed\_3c" feature map ($28\times 28\times 256$) is average-pooled to a 256-dimensional feature vector, which is used as the output of the base network.

\textbf{Baselines.} The baseline generalist model directly classifies all 6000 categories, whose architecture is the base network plus FC512-FC512 top layers. The generalist model serve as performance lower bound model. For specialists, we train five separate models, each model focuses on one vertical of categories, and the model architecture is the same as the generalist model. The specialists serve as performance upper bound model.

\textbf{Evaluation setup.} We train the models using Tensorflow by 50 GPU workers (P100) and 25 parameter severs. The batch size is set to be 64. Adagrad is used to optimized the networks, learning rate is set to 0.001. We use the same evaluation metric as in EFT, which is (mean) per-vertical average precision.
    
\textbf{Comparison on Openimage.} We compare three model variants with the baseline: FC512-FC512 with multi-teacher distillation (FC512-FC512+D), FC512-SC512 with multi-teacher distillation (FC512-SC512+D), FC512-SC512 with multi-teacher distillation and vertical-level self-paced learning (FC512-SC512+D+S). The results are shown in Table \ref{table:openimage}. Comparing FC512-FC512 and FC512-FC512+D, we can see that multi-teacher single-student knowledge distillation improves the performance from 62.0 to 63.2, which shows that multi-teacher distillation is able to effectively transferring the knowledge to the student model. Comparing FC512-SC512+D and FC512-FC512+D, the results show that the structurally connected layers expand the model capacity and thus improve the classification performance from 63.2 to 64.0. FC512-SC512+D+S improve the mean per-vertical average precision from 64.0 to 65.8 comparing with FC512-SC512+D, it can be seen that vertical level self-paced learning makes the student learn from different teachers more effectively.

It can be seen that the proposed multi-teacher distillation, self-paced learning and structurally connected layers consistently improve the student performance on both EFT and OpenImage. However, the performance gap between the student model and specialists on OpenImage is smaller than the gap on EFT. We think the main reason is that we use a much larger bottleneck layer (top-2) for the specialist (4096) than that for the generalist (512) on EFT, while on OpenImage, they are the same (512).

\section{Conclusion}
We tackle the problem of image classification at the scale of 100K classes. Training a single network at such a scale is challenging due to the intolerable large model size and slow training speed, and the performance is often unsatisfying. A straightforward solution would be training multiple specialists. However, deploying dozens of expert networks in a practical system significantly increase system complexity. We design a Knowledge Concentration method, which effectively merges the knowledge from dozens of specialists (\emph{i.e.} teachers) into a single student model. Specifically, we design a multi-teacher knowledge distillation framework and a self-paced learning mechanism to allow the student to learn from different teachers at various paces, and structurally connected layers to expand model capacity. Our method is evaluated on EFT and Open-Image datasets, and performs significantly better than the baseline model.

\textbf{Acknowledgements.}  This work would not have been possible without the efforts of members in Image Understanding group in Google Research. The authors would like to thank 
Lu Chen, Zhongli Ding and Zheyun Feng for helping in collecting and cleaning EFT dataset;
Tom Duerig and Futang Peng for previous internal works on knowledge distillation;
Neil Alldrin and Victor Gomes for OpenImage dataset collection and evaluation;
Danushen Gnanapragasam for the help in Flume pipeline;
Ariel Gordon for the TensorFlow training infra;  
Chen Sun, Xiangxin Zhu, Tom Duerig, and Chuck Rosenberg for reviewing the manuscript and providing valuable suggestions;
Howard Zhou, Lucy Gao, Pengchong Jin for the helpful discussions.

{\small
\bibliographystyle{ieee}
\bibliography{egbib}
}

\end{document}